\documentclass[runningheads]{llncs}
\usepackage{multirow}
\usepackage{geometry}
\usepackage[T1]{fontenc}
\usepackage{graphicx,verbatim}
\usepackage{booktabs} 
\usepackage{amsmath,amssymb,amsfonts}
\usepackage{caption}
\usepackage{url}
\usepackage{xcolor}
\usepackage[misc]{ifsym} 
\usepackage{bbding}

\begin{document}
\title{Uncertainty-Aware Concept and Motion Segmentation for Semi-Supervised Angiography Videos}
%

\author{Yu Luo\inst{1} \textsuperscript{*} \and
Guangyu Wei\inst{2,5} \textsuperscript{*} \and
Yangfan Li\inst{3} \textsuperscript{(\Letter)}  \and
Jieyu He\inst{4}\and
Yueming Lyu\inst{5}}

\authorrunning{Y. Luo et al.}

\institute{School of Mathematical Sciences, Ocean University of China, China \and
School of Haide, Ocean University of China, China \and
School of Computer Science and Engineering, Central South University, China \\ 
\email{liyangfan37@csu.edu.cn} \and
The Second Xiangya Hospital of Central South University, China \and
School of Intelligent Science and Technology, Nanjing University, China \\
}
  
\maketitle              

\renewcommand{\thefootnote}{\fnsymbol{footnote}}
\footnotetext[1]{These authors contributed equally to this work.}

\begin{abstract}
Segmentation of the main coronary artery from X-ray coronary angiography (XCA) sequences is crucial for the diagnosis of coronary artery diseases. However, this task is challenging due to issues such as blurred boundaries, inconsistent radiation contrast, complex motion patterns, and a lack of annotated images for training. Although Semi-Supervised Learning (SSL) can alleviate the annotation burden, conventional methods struggle with complicated temporal dynamics and unreliable uncertainty quantification.
To address these challenges, we propose SAM3-based Teacher-student framework with Motion-Aware consistency and Progressive Confidence Regularization (SMART), a semi-supervised vessel segmentation approach for X-ray angiography videos.
First, our method utilizes SAM3's unique promptable concept segmentation design and innovates a SAM3-based teacher-student framework to maximize the performance potential of both the teacher and the student. 
Second, we enhance segmentation by integrating the vessel mask warping technique and motion consistency loss to model complex vessel dynamics. To address the issue of unreliable teacher predictions caused by blurred boundaries and minimal contrast, we further propose a progressive confidence-aware consistency regularization to mitigate the risk of unreliable outputs.
Extensive experiments on three datasets of XCA sequences from different institutions demonstrate that SMART achieves state-of-the-art performance while requiring significantly fewer annotations, making it particularly valuable for real-world clinical applications where labeled data is scarce.
Our code is available at: \url{https://github.com/qimingfan10/SMART}.

\keywords{Semi-Supervised Learning \and Coronary Angiography \and Consistency regularization \and Promptable Concept Segmentation.}


\end{abstract}
\section{Introduction}

Coronary artery disease (CAD) remains the leading cause of morbidity and mortality worldwide, with the majority of cardiovascular disease deaths occurring in low- and middle-income countries and regions~\cite{1}. X\text{-}ray coronary angiography (XCA) is considered to the golden standard in clinical decision-making~\cite{2}. For the automatic diagnosis of vascular diseases, precise segmentation of coronary arteries from sequences of consecutive XCA images is essential~\cite{3}. 

Deep learning-based approaches for XCA sequences segmentation have shown promising results in recent years~\cite{24,25,26}. 
However, obtaining labeled data is extremely expensive and time\text{-}consuming in clinical scenarios, resulting in a significantly larger volume of unlabeled data compared to labeled samples. This constraint has sparked a surge of interest in semi\text{-}supervised learning (SSL) approaches~\cite{39,40,41}, which leverage extensive amounts of unlabeled data along with the limited labeled data to enhance model performance. 
Recently, the Segment Anything Model (SAM)~\cite{20} and its successor SAM2~\cite{21} have demonstrated impressive few-shot segmentation capabilities~\cite{23}, providing a promising approach to generate pseudo-labels in label‑scarce scenarios. However, SAM cannot be applied directly to medical image segmentation scenarios. To address this limitation, existing research has proposed the incorporation of precise prediction points, bounding boxes, or learnable feature prompts into SAM series to enhance the accuracy of pseudo‑labels for unlabeled data~\cite{22,27,28,13,14}.

Due to variations in X\text{-}ray imaging systems across different institutions, methods that rely on geometric cues or learnable feature prompts often struggle to generalize effectively across diverse clinical scenarios. Recently, innovative models~\cite{29,30,31} have begun to explore the integration of textual information to assist in segmentation tasks. Notably, SAM3~\cite{32} introduces an architectural shift by centering on concept prompts semantic phrases that denote visual objects. This design ensures that the system genuinely understands the semantics of visual structures. However, applying SAM3 directly to XCA sequences would ignore temporal dependencies and cross-modal alignment, potentially leading to temporally inconsistent segmentation and complicating the identification of vascular structures~\cite{33,34,35}. These challenges are further exacerbated by inherent limitations of XCA, including significant temporal discontinuities in target morphology and scale due to involuntary organ motion, as well as minimal contrast and low signal-to-noise ratios~\cite{36,37,38}. 

To address the challenges mentioned earlier, we propose SMART (\textbf{S}A\textbf{M}3-Based Uncertainty-\textbf{A}ware Confidence \textbf{R}egularization with Motion Consistency for \textbf{T}eacher-Student Architecture), a semi-supervised mean-teacher approach for vessel segmentation in X\text{-}ray videos. We first eliminate the dependency on geometric prompts by leveraging the promptable concept segmentation ability of SAM3, which is guided by textual descriptions, allowing for capturing localized details and complex boundaries with improved accuracy. 
Considering the minimal contrast of coronary arteries and motion-induced blurring, the generation of pseudo-labels may be inherently unreliable and noisy. To address this issue, we design a novel progressive confidence-aware consistency regularization. This approach enables the framework to dynamically adjust the utilization of supervisory signals from unlabeled data across different training stages, while enhancing the invariance of predictions. Consequently, it ensures more reliable and stable segmentation results even when learning from less dependable pseudo‑labels.
Recognizing the unique characteristics of cardiac movements and vessel flow patterns, we incorporate mask warping operation, motion consistency loss and flow coherence loss, which ensures temporally consistent and detailed segmentation while effectively distinguishing vascular foreground from background.
Our method has been extensively evaluated on three X\text{-}ray coronary angiography video datasets, demonstrating superior performance over existing methods, especially when labeled data are extremely limited. Specifically, using only 16 labeled videos, each with only one or two annotated frames, our method achieves an improvement of over 6\% Dice compared to various strong baselines.

\section{Method} 

\begin{figure}[t]
    \centering
    \includegraphics[width=0.9\textwidth]{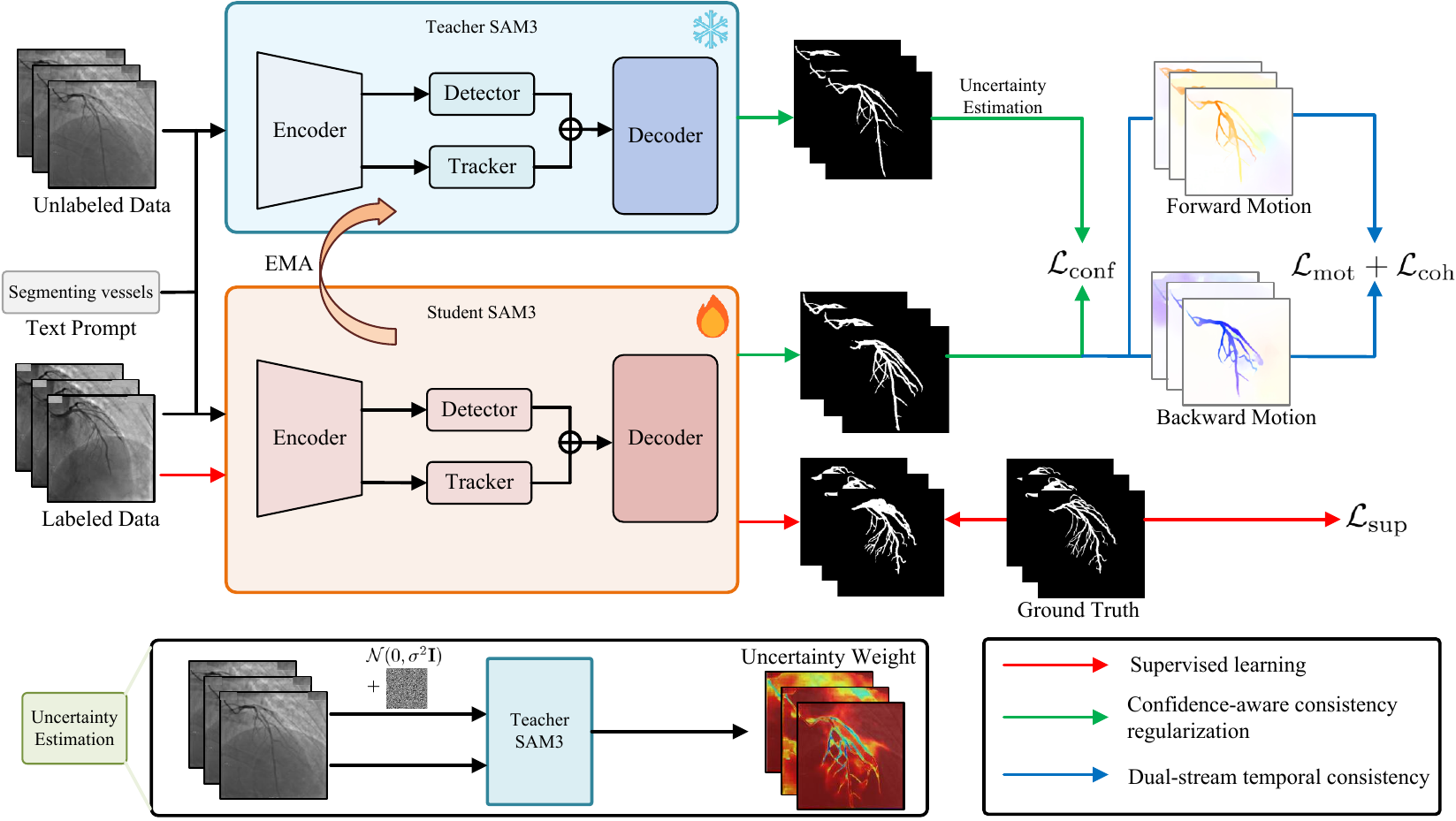}
    \caption{The overview of our proposed method. SMART (SAM3-based Motion-Aware Confidence Regularization for Teacher-Student Architecture) model enhances the reliability of teacher model outputs through a confidence-aware consistency loss \(\mathcal{L}_{\text{conf}}\). It incorporates temporal understanding from video sequences via dual-stream temporal consistency losses \(\mathcal{L}_{\text{opti}}\) and \(\mathcal{L}_{\text{coh}}\), while effectively leveraging annotated data through the supervised loss \(\mathcal{L}_{\text{sup}}\).}
    \label{fig:framework}
\end{figure}

\textbf{Overview.} 
The goal of SSL segmentation is to approach the performance of a fully supervised model using a limited amount of labeled data $D_l = { (\mathbf{x}_i, \mathbf{y}_i) }_{i=1}^N $ and a large quantity of unlabeled data $ D_u = { (\mathbf{x}_i) }_{i=N+1}^{N+M} $, where $ N \ll M $. Here, $ \mathbf{x}_i \in \mathbb{R}^{H \times W} $ represents an $ H \times W $ image, and $ \mathbf{y}_i \in \{0, 1\}^{H \times W} $ denotes the corresponding annotation for the labeled data.
Figure~\ref{fig:framework} provides an overview of the SMART architecture, which employs a teacher model $f_{\Theta_T}$ and a student model $f_{\Theta_S}$, both utilizing the text prompt for segmentation. The training process consists of two sequential stages: \textit{text-driven segmentation fine-tuning} and \textit{motion-aware semi-supervised learning}. Specifically, in the first stage, we apply a fine-tuning strategy to the teacher model $f_{\Theta_T}$ on labeled data $D_l$, preserving generic visual and language features while incorporating medical domain knowledge. In the second stage, the fine-tuned teacher is frozen and guides the student on unlabeled data $D_{u}$. Crucially, we propose a dual-stream temporal consistency strategy and motion consistency loss and a confidence-aware consistency regularization to enhance the student model's effective utilization of temporal information in video sequences and its robustness to unreliable teacher outputs. During the inference stage, we use only the student model $f_{\Theta_S}$ for predictions.

\subsection{Text-driven Segmentation Fine-tuning}

The utilization of unlabeled data is a critical component of SSL, where pseudo-labels generated by the teacher provide essential guidance for the student to learn effectively from limited annotations. Although SAM3 has achieved remarkable success in the natural image domain, it lacks an understanding of domain-specific semantics in the medical field~\cite{35}. 
In the first stage, we address this requirement using a foundational technique: transferring the concept of visual instruction tuning~\cite{45} to SAM3. By leveraging labeled data for initial optimization, this step enables the teacher model to adapt more effectively to the target task. The teacher SAM3 retains its original architecture, with training applied solely to the text prompts. Specifically, we fine-tune the image encoder, text encoder, and detector of SAM3 using diverse prompts related to vessel segmentation, while keeping other components frozen. During training, these text prompts are optimized to segment the target objects of interest. The finetuning optimization is formulated as:
\begin{equation}
    \mathcal{L}_{\text{ft}}(p(\mathbf{x}_i^l),\mathbf{y}_i^l)=\lambda_1 \mathcal{L}_{\text{Dice}}(p(\mathbf{x}_i^l),\mathbf{y}_i^l)+\lambda_2\mathcal{L}_{\text{Bce}}(p(\mathbf{x}_i^l),\mathbf{y}_i^l),
\end{equation}
wherep $p(\mathbf{x}_i^l)=f_{\Theta_T}(\mathbf{x}_i^l))$ denotes the teacher SAM3's prediction for a labeled image, and $\mathbf{y}_i^l$ is the corresponding ground-truth mask. $\mathcal{L}_{\text{Dice}}$ and $\mathcal{L}_{\text{Bce}}$ represent the Dice loss and Cross-entropy loss, respectively.

\subsection{Confidence-aware Consistency Regularization}

The utilization of pseudo-labels generated by the teacher addresses the challenge of leveraging unlabeled data to enhance semi-supervised learning performance. The success of this approach is generating reliable predictions for unlabeled data. However, the inherent limitations of coronary angiography images, such as low signal-to-noise ratios and poor contrast, can lead to unreliable outputs from SAM3, characterized by low accuracy and high variance in low-quality regions. To mitigate this issue, it is crucial to enhance output consistency under varying noise conditions and to adaptively down-weight uncertain regions at different stages of training. In the ideal case, predictions for target vessels should remain stable across different noise perturbations, while the model progressively strengthens its learning on uncertain regions (e.g., low-contrast distal vessels) as training proceeds. Motivated by these considerations, we propose a novel Confidence-aware Consistency Regularization. Specifically, the teacher first receives multiple noise-perturbed versions of the video frames \(\mathbf{X} \in \mathbb{R}^{C \times H \times W}\). The perturbations are sampled as \(\boldsymbol{\epsilon}^{(i)} \sim \mathcal{N}(0, \sigma^2\mathbf{I})\), where \(i=1,\dots,N\), generating \(N\) independent predictions denoted as \(\mathcal{P} = \{\hat{p}_i\}_{i=1}^N=\{f_{\Theta_T}(\mathbf{X}+\mathbf{\epsilon}^{(i)})\}_{i=1}^N\). This noise injection simulates the quality variations commonly observed in coronary angiography images. 
Under these perturbations, the predictions \(\hat{p}_i\) are expected to remain consistent across different instances of noise.
Since an ensemble of multiple predictions typically yields more robust results, we adopt the averaged prediction \(\bar{\mathbf{P}}={1}/{N} \sum_{i=1}^{N} \hat{p}_i\) as reliable guidance for the student. 

Meanwhile, to ensure that the model enhances supervision in regions with high uncertainty, we compute the uncertainty weight as $\boldsymbol{\mathcal{U}} = 1/{N} \sum_{i=1}^{N} \left( f_{\Theta_T}(\mathbf{X}^{(i)}) - \bar{\mathbf{P}} \right)^2$. Subsequently, the basic consistency distance is derived from the student output $\mathbf{S} = f_{\Theta_S}(\mathbf{X})$ and the averaged prediction $\bar{\mathbf{P}}$, which can be expressed as:
\begin{equation}
  \mathcal{D}(x,y)=\left(\sigma(S(x,y))-\sigma(\bar{P}(x,y))\right)^2,  
\end{equation}
where $\sigma(\cdot)$ represents the sigmoid function. By weighting the basic consistency distance with the derived uncertainty weight, SMART achieves adjustable supervision intensity, assigning higher importance to regions with elevated uncertainty to drive improvement. The confidence-aware consistency loss can then be formulated as:
\begin{equation}
\mathcal{L}_{\text{conf}} = \frac{\displaystyle \sum_{x=1}^{H} \sum_{y=1}^{W} \mathcal{D}(x,y) \mathcal{U}(x,y)}
{\displaystyle \sum_{x=1}^{H} \sum_{y=1}^{W} \mathcal{U}(x,y) + N\eta}
+ \frac{\beta}{N} \sum_{x=1}^{H} \sum_{y=1}^{W} \mathcal{U}(x,y),
\label{eq:confidence_loss_single_simplified}
\end{equation}
where \(\beta\) is the uncertainty regularization weight, and \(\eta\) is a small stability factor (e.g., \(10^{-6}\)) to avoid division by zero.

\subsection{Dual-Stream Temporal Consistency}
Compared to static medical images, XCA videos exhibit more complex temporal dynamics. To enhance the temporal continuity of vessel segmentation and prevent sudden changes in predictions between adjacent frames, we propose a dual‑stream temporal consistency strategy based on optical flow. This approach effectively leverages the spatial-temporal information inherent in video sequences.

To mitigate the confirmation bias associated with unidirectional flow in motion modeling, we adopt a dual‑stream estimation strategy. Specifically, we utilize a pretrained optical flow estimator, SEA\text{-}RAFT~\cite{44}, to compute both the forward flow \(\mathbf{F}_{t \to t+1}\) and the backward flow \(\mathbf{F}_{t+1 \to t}\) for consecutive frames \(\mathbf{X}_t\) and \(\mathbf{X}_{t+1}\).
Due to the limited number of frames in current XCA videos~\cite{42}, the topological structure of blood vessels remains unchanged during this short period. Based on the estimated optical flow, we introduce a motion consistency loss function to ensure the temporal coherence of the predicted masks across consecutive frames.This process can be formulated as:
\begin{equation}
    \mathcal{L}_{\text{opti}} = \frac{1}{2N} \sum_{x=1}^{H} \sum_{y=1}^{W} \left( \left[ \mathbf{S}_t(x,y) - \mathcal{W}(\mathbf{S}_{t+1}, \mathbf{F}_{t\to t+1})(x,y) \right]^2 + \left[ \mathbf{S}_{t+1}(x,y) - \mathcal{W}(\mathbf{S}_t, \mathbf{F}_{t+1\to t})(x,y) \right]^2 \right),
    \label{eq:motion_consistency_loss}
\end{equation}
where $\mathcal{W}(\mathbf{S}, \mathbf{F})(x,y) = \mathbf{S}\big( x + \mathbf{F}_u(x,y),\; y + \mathbf{F}_v(x,y) \big)$ denotes the mask warping operation.

While the motion consistency loss ensures pixel\text{-}level temporal alignment, it still faces inherent ambiguity at vascular boundaries. To address this issue, we propose a flow coherence loss that enables the model to distinguish the foreground from the background based on differences in motion patterns. This approach penalizes deviations of boundary points from the dominant motion of the vessel body. This process can be expressed as:
\begin{equation}
\begin{cases}
    \bar{\Phi} =  
\frac{\displaystyle \sum_{x=1}^{W} \sum_{y=1}^{H} \hat{S_t}(x,y) \cdot F_{t\rightarrow t+1}(x,y)}
{\displaystyle \sum_{x=1}^{W} \sum_{y=1}^{H} \hat{S_t}(x,y) + \epsilon},\\[1.5ex]
    \mathcal{L}_{\text{coh}} = \dfrac{\lambda_{\text{coh}}}{\displaystyle\sum_{x=1}^{H} \sum_{y=1}^{W} \hat{S_t}(x,y)} \displaystyle\sum_{x=1}^{H} \sum_{y=1}^{W} \hat{S_t}(x,y) \left\| \mathbf{F}_{t\rightarrow t+1}(x,y) - \bar{\mathbf{\Phi}} \right\|_2^2 ,\label{eq:flow_grouping_loss}
\end{cases}
\end{equation}
where $\hat{S_t}(x,y)$ represents the student prediction mask for frame t.

In this study, the supervised segmentation loss consists of the Dice and cross-entropy (CE) losses, which can be defined by $\mathcal{L}_{\text{seg}}(y,\hat{y})=\lambda_{\text{Dice}}\mathcal{L}_{\text{Dice}}(y,\hat{y})+\lambda_{\text{Bce}}\mathcal{L}_{\text{Bce}}(y,\hat{y})$, where $\hat{y}$ and $y$ denote the prediction and ground truth, respectively.
In summary, the overall optimization objective for semi-supervised learning is written as:
\begin{equation}
\mathcal{L}_{\text{all}}=\lambda_{\text{Dice}}\mathcal{L}_{\text{Dice}}+\lambda_{\text{Bce}}\mathcal{L}_{\text{Bce}}+\lambda_{\text{conf}}\mathcal{L}_{\text{conf}}+\lambda_{\text{opti}}\mathcal{L}_{\text{opti}}+\lambda_{\text{coh}}\mathcal{L}_{\text{coh}}.
\end{equation}

\begin{table*}[h]
\centering
\caption{Quantitative results on the XCAV and CAVSA datasets.}
\setlength{\tabcolsep}{3pt} 
\resizebox{\textwidth}{!}{
\begin{tabular}{l|c|c|c|c|c|c||c|c|c|c|c|c}
\toprule
\multirow{2}{*}{Method} & \multicolumn{6}{c||}{\textbf{XCAV}} & \multicolumn{6}{c}{\textbf{CAVSA}} \\
\cline{2-13}
 &  \#Lab & DSC$\uparrow$ & NSD$\uparrow$ & clDice$\uparrow$ & Spe$\uparrow$ & Sen$\uparrow$ & \#Lab & DSC$\uparrow$ & NSD$\uparrow$ & clDice$\uparrow$ & Spe$\uparrow$ & Sen$\uparrow$ \\
\midrule
UNet~\cite{17}  & 16/111 & 70.80 & 60.12 & 69.24 & 97.75 & 66.33 & 16/1061 & 64.19 & 54.25 & 70.27 & \textbf{99.08} & 53.36 \\
MedSAM2~\cite{22} & 16/111 & 22.37 & 21.07 & 3.57 & 84.14 & 32.01 & 16/1061 &33.13 & 22.05 & 33.03 & 86.30 & 44.31 \\
SAM3~\cite{32} & 16/111 & 42.73 & 30.86 & 34.51 & 98.11 & 35.59 & 16/1061 &30.82 & 25.25 & 30.14 & 92.99 & 30.18 \\
Denver~\cite{42} & 16/111 & 73.30 & 51.50 & 70.40 & 98.50 & 65.60 & 16/1061 &76.53 & 68.39 & 79.17 & 98.90 & 75.28 \\
KnowSAM~\cite{13} & 16/111 & 72.76 & 77.67 & 77.08 & 98.07 & 66.77 & 16/1061 &72.76 & 60.57 & 76.12 & 98.07 & 66.77 \\
CPC-SAM~\cite{14} & 16/111 & 77.90 & 81.97 & 79.15 & \textbf{98.53} & 72.09 & 16/1061 &77.90 & 65.34 & 78.28 & 98.53 & 72.09 \\
\textbf{SMART (Ours)} & 16/111 & \textbf{84.39} & \textbf{89.13} & \textbf{83.01} & 86.25 & \textbf{81.83}  & 16/1061 &\textbf{91.00} & \textbf{91.71} & \textbf{97.73} & 91.70 & \textbf{96.30} \\
\bottomrule
\end{tabular}
}

\label{tab:main_result}
\end{table*}
\begin{figure}[ht]
    \centering
    \includegraphics[width=\textwidth]{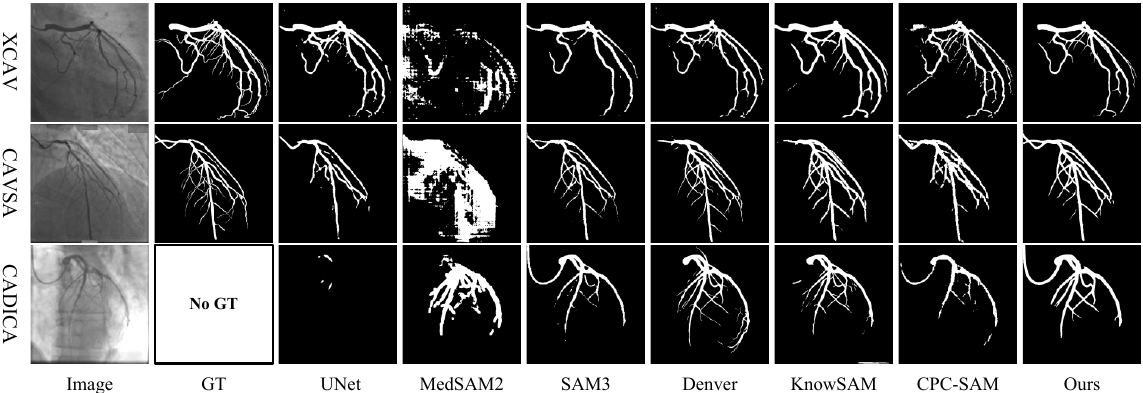}
    \caption{Visualization results on the vessel segmentation.}
    \label{fig:main_result}
\end{figure}
\section{Experiments}
\subsection{Experimental Setup}
\subsubsection{Datasets.}
We evaluate our proposed method on two publicly available datasets, the XCAV dataset~\cite{42} and the CADICA dataset~\cite{43}, as well as one private dataset, referred to as the CAVSA (Coronary Angiography Video Segmentation) dataset. The XCAV dataset includes 111 complete X-ray coronary angiography video recordings from 59 patients, with one or two frames per video annotated at the point of maximal contrast agent prominence, whereas the CADICA dataset includes 662 complete recordings from 42 patients without corresponding ground‑truth annotations, which are used to evaluate generalization performance.
We collect 1,061 complete coronary X-ray angiography video sequences from 121 patients.
Each sequence contains an average of 79 frames, each with a high resolution of $512 \times 512$ pixels. We invited experienced radiologists to annotate the vascular regions, providing complete annotations for 16 video sequences and annotating 1 to 2 frames per additional sequence at the point of maximal contrast agent prominence. Following prior work, we split the XCAV and CAVSA datasets into training and test sets with an 8:2 ratio.
\begin{table*}[h]
  \centering
  \scriptsize
  \renewcommand{\arraystretch}{1.15}
  \setlength\tabcolsep{2.8pt}
  \caption{Ablation study on the key components.}
  \label{tab:tab_Ablation}
\begin{tabular}{ccc|ccccc|ccccc}
\hline
\multicolumn{3}{c|}{Settings} & \multicolumn{5}{c|}{\textbf{XCAV}} & \multicolumn{5}{c}{\textbf{CAVSA}}\\
\hline
TPT & CCR     & DSTC     & DSC$\uparrow$   & NSD~$\uparrow$  & clDice~$\uparrow$  & Spe~$\uparrow$ & Sen~$\uparrow$  & DSC$\uparrow$   & NSD~$\uparrow$  & clDice~$\uparrow$  & Spe~$\uparrow$ & Sen~$\uparrow$  \\
\hline
 & \checkmark  & \checkmark        &   82.38   & 83.42  & 79.84 & \textbf{99.16} &  77.26  &  78.87 &  67.35 &  81.17  & 99.42 & 73.80 \\
 \checkmark & & \checkmark  &   76.71   & 78.29 & 79.86    &  97.46 &   78.69   &   25.82 &  21.24 &  32.65  & \textbf{99.77} & 19.38  \\
 \checkmark & \checkmark &   & 76.24   & 77.27   & 78.53 & 98.05  & 74.33      &  47.77 & 40.71  &  50.37   & 99.59     &  38.46 \\
\checkmark  &\checkmark  &  \checkmark  & \textbf{84.39} & \textbf{89.13} & \textbf{83.01} &  86.25 &  \textbf{81.83}  & \textbf{91.00} & \textbf{91.71} & \textbf{97.73} &  91.70 & \textbf{96.30}  \\
\hline
\end{tabular}
\end{table*}

\subsubsection{Implementation Details and Evaluation Metrics}
SMART is implemented by PyTorch and trained on an NVIDIA A40 GPU. We utilize the AdamW\cite{LoshchilovH19} optimizer with an initial learning rate of 1e-4 and weight decay of 0.01. The batch size is set to 4 and the numer of iterations is set to 6k. Then, we gradually incorporate $\mathcal{L}_{\text{Dice}}\; (\lambda_{\text{1}}=0.05)$, $\mathcal{L}_{\text{Bce}}\; (\lambda_{\text{2}}=0.95)$, $\mathcal{L}_{\text{Dice}}\; (\lambda_{\text{Dice}}=0.5)$, $\mathcal{L}_{\text{Bce}}\; (\lambda_{\text{Bce}}=0.5)$, $\mathcal{L}_{\text{conf}}\; (\lambda_{\text{conf}}=0.5)$, $\mathcal{L}_{\text{opti}}\; (\lambda_{\text{opti}}=0.3)$ and $\mathcal{L}_{\text{coh}}\; (\lambda_{\text{coh}}=0.2)$.
The number of noise perturbations applied to each image for uncertainty estimation is set to \(N = 8\).
We apply different data augmentation strategies to the teacher and student models. The student receives weak augmentations: rotation (\(\pm 5^\circ\)), scaling (\(0.95\)–\(1.05\)), and low noise (\(\sigma = 0.01\)). In contrast, the teacher is subjected to stronger augmentations: rotation (\(\pm 15^\circ\)), scaling (\(0.85\)–\(1.15\)), and higher noise (\(\sigma = 0.03\)).
In the preprocessing stage, we compute the optical flow using SEA-RAFT~\cite{44}
Following the prior work~\cite{42}, five evaluation metrics are taken, including the dice similariy coefficient (DSC), normal surface distance (NSD), centrelineDice (clDice), specificity (Spe), sensitivity (Sen).

\begin{figure}[t]
    \centering
    \includegraphics[width=\textwidth]{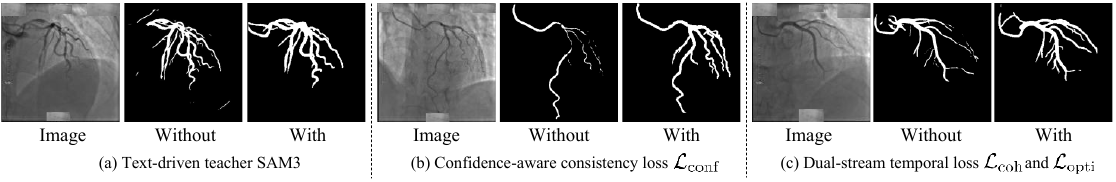}
    \caption{Visual comparisons of ablation studies.}
    \label{fig:ablation}
\end{figure}

\subsection{Comparisons with the State-of-the-arts.} We compare our method with state-of-the-art (SOTA) SAM-based SSL methods, including KnowSAM~\cite{13}, CPC-SAM~\cite{14}. Additionally, we compare with the supervised counterparts trained with labeled data alone, i.e, UNet~\cite{17}, MedSAM2~\cite{22}, and SAM3~\cite{32}. Furthermore, we conduct experiments using Denver~\cite{42}, a method specifically designed for XCA video segmentation.
Table~\ref{tab:main_result} presents a quantitative comparison and experiment results on the XCAV dataset clearly demonstrate the superiority of out proposed SMART method compared to several methods. With only 14\% labeled video, SMART achieves a DSC of 84.39\% and a clDice of 83.01\%, outperforming the next best method, CPC-SAM, by 6.49\% and 3.86\%, respectively. Using only 1.5\% of the labeled data, our method achieves a 13.1\% improvement in DSC on the CAVSA dataset. The visualization comparisons in Fig.~\ref{fig:main_result} further validate the superiority of our method. 

We further evaluate the generalization ability of SMART and competing methods on the CADICA dataset, as illustrated in Fig.~\ref{fig:main_result}. Purely supervised methods are constrained by the domain of training data and thus struggle to generalize effectively. SAM-based semi-supervised approaches, which rely on geometric priors or learnable feature prompts, also exhibit limited generalization performance. In contrast, SMART leverages the promptable concept segmentation capability of SAM3, demonstrating superior flexibility and generalization across diverse clinical scenarios.

\begin{table}[t]
\centering
\caption{Effects of different numbers of noise perturbations on the XCAV dataset.}
\label{tab:number}
\scriptsize
\begin{tabular}{c c c c c c}
\toprule
\ Noise & DSC(\%)$\uparrow$ & NSD(\%)$\uparrow$ & clDice(\%)$\uparrow$ & Spe(\%)$\uparrow$ & Sen(\%)$\uparrow$ \\
\midrule
2  & 83.59 & 88.45 & 81.82 & 85.38 & 79.91 \\
4  & 83.73 & 87.53 & 82.02 & 85.14 & 81.76 \\
6  & 83.79 & 88.40 & 82.51 & 85.47 & 79.41 \\
8  & \textbf{84.39} & \textbf{89.13} & \textbf{83.01} & \textbf{86.25} & \textbf{81.83} \\
\bottomrule
\end{tabular}
\end{table}

\subsection{Ablation Studies}

\textbf{Effectiveness of Text-driven Segmentation Fine-tuning.}
We apply fine-tuning to the teacher model prior to semi-supervised learning. To demonstrate the effectiveness of promptable segmentation tuning, we compare our method with the point‑prompted SAM3 baseline, which extracts center points along the vascular skeleton at regular intervals as geometric prior prompts. As shown in Table~\ref{tab:tab_Ablation} and Fig.~\ref{fig:ablation} (a), the Dice coefficients improve from 82.38\% and 78.87\% to 84.39\% and 91.00\% on the XCAV and CAVSA datasets, respectively.

\noindent  \textbf{Effectiveness of Confidence-aware Consistency Regularization.} 
We conduct an experiment to assess the effect of the confidence-aware consistency loss by removing it from the training pipeline. As shown in Table~\ref{tab:tab_Ablation}, the segmentation performance drops significantly, with the Dice score decreasing by 43.23\%. This indicates that without regularizing the teacher SAM3's outputs, the segmentation results are adversely affected. The comparison in Fig.~\ref{fig:ablation} (b) further demonstrates the effectiveness of this regularization term.

\noindent \textbf{Effectiveness of Dual-Stream Temporal Consistency.} 
We further validate the efficacy of the dual‑stream temporal consistency losses \(\mathcal{L}_{\text{opti}}\) and \(\mathcal{L}_{\text{coh}}\) in Table~\ref{tab:tab_Ablation}. Specifically, the clDice score improves by approximately 39\%, indicating a substantial gain in the spatial connectivity of segmentation. The results clearly demonstrate that leveraging temporal information from video effectively enhances segmentation continuity and helps avoid disconnected or over-segmented predictions.

\noindent \textbf{Effect of the Number of Noise Perturbations.} Table~\ref{tab:number} analyzes the impact of different numbers of noise perturbations on SMART, evaluating their contribution to the confidence-aware consistency regularization. The results indicate that the number of perturbations plays a critical role: a higher number enables a more accurate estimation of uncertainty in the teacher's outputs, which directly affects the evaluation metrics. Specifically, the clDice coefficient improves from 81.82\% to 83.01\% on the XCAV dataset as the number of perturbations increases.

\section{Conclusion}

In this work, we develop a SAM3-based teacher-student framework for semi-supervised medical image segmentation, utilizing the promptable concept segmentation capabilities of SAM3. First, we implement a vision instruction tuning strategy to align the anatomical semantics of the teacher's encoder with the segmentation masks generated by the decoder through fine-tuning, thereby providing effective guidance to the student SAM3. Next, our framework incorporates confidence-aware consistency and dual-stream temporal consistency to address the unreliability of teacher outputs and leverage motion information in video sequences for improved segmentation performance. Through comparative experiments on three datasets, we validate the effectiveness and generalizability of the proposed method, demonstrating its potential to enhance clinical applicability.

%
%
%
\newpage
\bibliographystyle{splncs04}
\bibliography{mybibliography}

\end{document}